\documentclass[sigconf]{acmart}

\usepackage{subcaption}

\usepackage{algorithm, amsmath}
\usepackage{algorithmic}

\AtBeginDocument{%
  \providecommand\BibTeX{{%
    \normalfont B\kern-0.5em{\scshape i\kern-0.25em b}\kern-0.8em\TeX}}}
    
\setcopyright{none}
\renewcommand\footnotetextcopyrightpermission[1]{} 
\begin{document}

\title{FaIRCoP: Facial Image Retrieval using Contrastive Personalization}



\author{Devansh Gupta*}
\email{devansh19160@iiitd.ac.in}
\affiliation{%
  \institution{Indraprastha Institute of Information Technology, Delhi}
  \country{India}
}

\author{Aditya Saini*}
\affiliation{%
  \institution{Indraprastha Institute of Information Technology, Delhi}
  \country{India}
 }
\email{aditya18125@iiitd.ac.in}

\author{Drishti Bhasin*}
\email{drishti_b@me.iitr.ac.in}
\affiliation{%
  \institution{Indian Institute of Technology, Roorkee}
  \country{India}
}

\author{Sarthak Bhagat}
\email{sarthak16189@iiitd.ac.in}
\affiliation{%
 \institution{Indraprastha Institute of Information Technology, Delhi}
 \country{India}
}

\author{Shagun Uppal}
\email{shagun16088@iiitd.ac.in}
\affiliation{%
 \institution{Indraprastha Institute of Information Technology, Delhi}
 \country{India}
}

\author{Rishi Raj Jain}
\email{rishi19304@iiitd.ac.in}
\affiliation{%
  \institution{Indraprastha Institute of Information Technology, Delhi}
  \country{India}
 }
 
 \author{Ponnurangam Kumaraguru}
\email{pk.guru@iiit.ac.in}
\affiliation{%
  \institution{International Institute of Information Technology, Hyderabad}
  \country{India}
 }
 
  \author{Rajiv Ratn Shah}
\email{rajivratn@iiitd.ac.in}
\affiliation{%
  \institution{Indraprastha Institute of Information Technology, Delhi}
  \country{India}
 }





\begin{abstract}
  Retrieving facial images from attributes plays a vital role in various systems such as face recognition and suspect identification. Compared to other image retrieval tasks, facial image retrieval is more challenging due to the high subjectivity involved in describing a person’s facial features. Existing methods do so by comparing specific characteristics from the user’s mental image against the suggested images via high-level supervision such as using natural language. In contrast, we propose a method that uses a relatively simpler form of binary supervision by utilizing the user's feedback to label images as either similar or dissimilar to the target image. Such supervision enables us to exploit the contrastive learning paradigm for encapsulating each user's personalized notion of similarity. For this, we propose a novel loss function optimized online via user feedback.
  We validate the efficacy of our proposed approach using a carefully designed testbed to simulate user feedback and a large-scale user study. Our experiments demonstrate that our method iteratively improves personalization, leading to faster convergence and enhanced recommendation relevance, thereby, improving user satisfaction. Our proposed framework is also equipped with a user-friendly web interface with a real-time experience for facial image retrieval.
\end{abstract}





\maketitle
\pagestyle{plain}

\section{Introduction}

Facial image retrieval plays an important role in the domain of digital forensics for several tasks such as facial recognition \cite{conf/cvpr/SchroffKP15} and suspect identification \cite{Shrivastava2019FacefetchAE,JainSeekSuspect}. These systems aim to retrieve the images most similar to the query image by narrowing down the search space using image attribute descriptions. Such supervision can be provided in the form of detailed natural language descriptions \cite{g2g} which can be expensive to annotate as well as error prone, especially in tasks such as suspect retrieval where the witness often relies only on their visual memory. Moreover, the high degrees of variation in attributes including pose, illumination, expressions, and occlusions present in different facial images adds to the challenge of efficiently developing such systems. In this work, we address these challenges by developing a weakly-supervised facial image retrieval system with efficient real-time usage. For this, we utilize high-level categorical feature attributes as a weaker form of supervision capturing the user's notion of similarity (or dissimilarity) and propose a novel relevance feedback mechanism by incorporating these cues.


\begin{figure*}[h]
\begin{subfigure}{0.60\textwidth}
    \centering
    \includegraphics[scale=0.32]{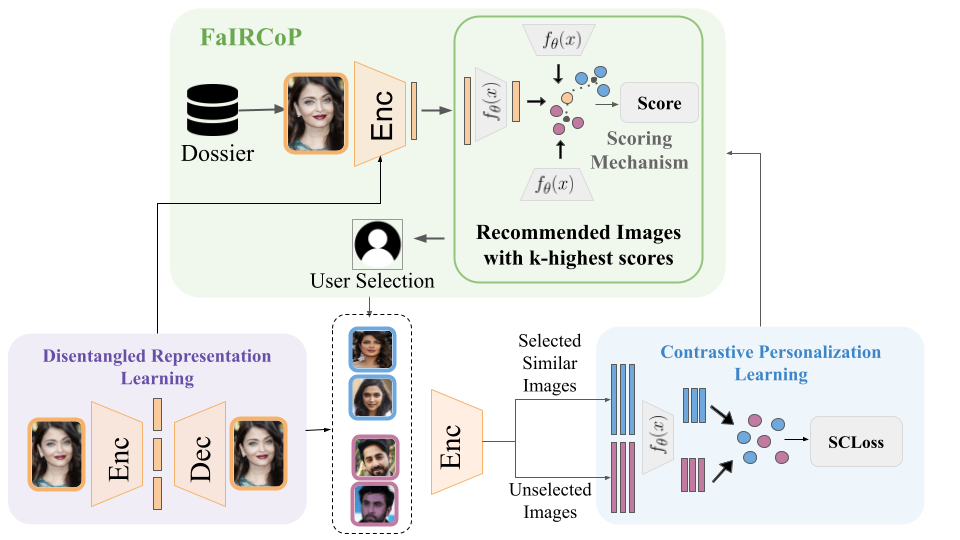}
    \caption{Illustration of our proposed framework ${\tt FaIRCoP}$ for facial image retrieval. We define the SCLoss for contrastive personalization learning in Equation~\ref{loss}.}
    \label{fig:FaIRCoP}
\end{subfigure}
\hfill
\begin{subfigure}{0.36\textwidth}
    \centering
    \includegraphics[scale=0.35]{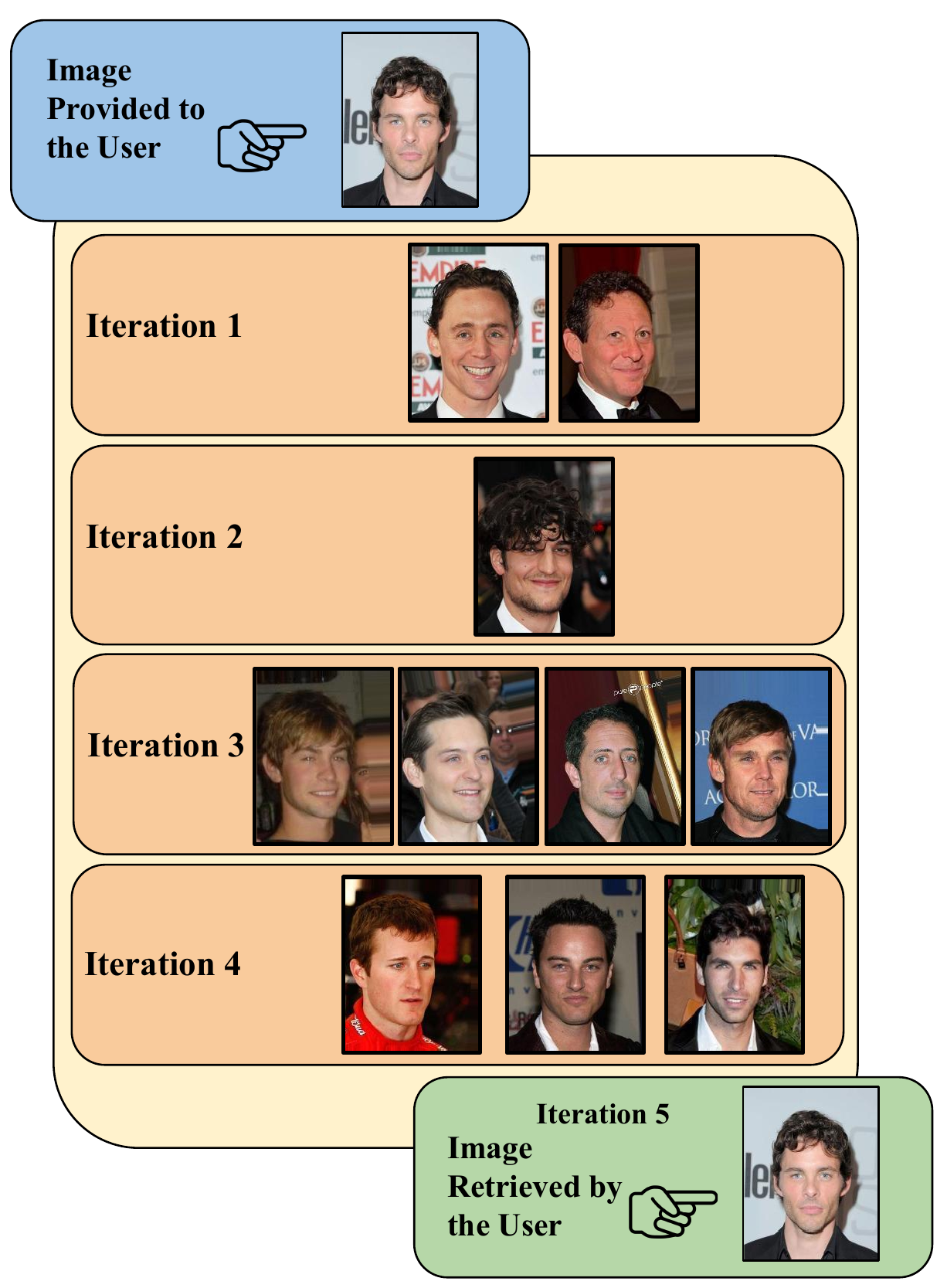}
    \caption{A search result from user study on our system showing similar images selected at each iteration.
    }
    \label{fig:faircop_p}
\end{subfigure}
\caption{${\tt FaIRCoP}$ - Facial Image Retrieval using Contrastive Personalization system. For security purposes, we do not show the images from the Criminal dataset and only use the CelebA dataset for demonstration.}
\end{figure*}


Prior work in the area of facial image retrieval has focused on the utilization of predefined annotated features to retrieve images from a database \cite{AttrMani,g2g}. This approach limits the user's expressibility to a limited number of tangible attributes and is expensive to train due to the requirement of feature annotations. To alleviate this issue, user feedback has been used to obtain relevant images in an online manner. In such cases, facial features become subject to the user's interpretation, making it crucial for the system to appropriately model user preferences. \citet{g2g,AttrMani,shapemani} utilized user feedback which explicitly mentions the changes as a query and suggested images. These approaches tend to impose a higher cognitive load on the user since they require them to recall the image's fine details from their visual memory. More user-friendly approaches such as \cite{sgdfi,Bhattarai_2016_CVPR,rocchio} successfully diminish the cognitive load by requiring the user to classify the mental image based on certain predefined parameters. \cite{JainSeekSuspect} exploited the similarity-based user feedback mechanism to learn the notion of divergence between similar and dissimilar images with respect to the user's mental image model. Despite learning representations aiming to personalise user preferences. These approaches failed to learn a distance metric space that encapsulates the variability among various factors of variation within the image. 

In this work, we propose a contrastive learning framework for facial image retrieval that adapts to each user's personalized notion of similarity (or dissimilarity). Our novel loss function, called the Separating Cluster loss, clusters the similar images as marked by the user while establishing the dichotomy between the similar and the dissimilar images. We also utilize unsupervised disentangled representation learning to obtain robust image embeddings which separate multiple facial attributes into partitioned latent spaces. This makes the representations more interpretable and also aid in efficiently narrowing the search space without the explicit dependency on labels, which can be expensive as well as noisy. Since our approach requires human interaction as part of our pipeline, our proposed algorithm referred to as the \emph{Facial Image Retrieval using Contrastive Personalization ${\tt( FaIRCoP}$)}, is equipped with a user-friendly web-based interface for retrieving images in real-time. \footnote{We provide more details of the web interface in the supplementary.}We also designed a custom user simulator with simulated human feedback to compare our method against various baselines and design choices before performing an extensive user study on two extensive facial image datasets. Figure \ref{fig:FaIRCoP} illustrates the pipeline of our proposed framework. 

We summarize our contributions as follows:

\begin{itemize}
    \item A novel contrastive learning-based loss function called the \textit{Separating Cluster Loss} for iteratively modifying the search space by clustering similar and dissimilar images.
    \item A relevance feedback framework, referred to as ${\tt FaIRCoP}$, which uses the above mentioned loss for incorporating personalized user feedback.
    \item A custom simulator to automate the user feedback for the proposed image retrieval method.
    \item A responsive web-based interface for real-time facial image retrieval equipped with our proposed algorithm.
\end{itemize}

\section{Related Works}

\paragraph{Disentangled Representation Learning.}

Disentangled representation learning is an approach for encoding high-dimensional data into independent low-dimensional latent space partitions, each capturing a distinct factor of variation. Several works \cite{Jha2018DisentanglingFO,Szab2018ChallengesID,Mathieu2016DisentanglingFO} followed this by exploiting limited supervision in order to extract the specified attribute from the rest of the underlying factors of variations. The resulting embeddings provide the model with enhanced interpretability and downstream task performance. However, they are subjected to inherent biases due to their dependence on specified feature annotations of single or multiple factors. Due to these reasons, unsupervised disentangled representation learning has gained traction in the community. Various prior works \cite{Hu2018DisentanglingFO,Shukla2019PrOSePO,Bhagat2020DisentanglingMF} focus on learning factored representations in a completely unsupervised manner. Such representations that capture each tangible feature into a discrete chunk within the latent space compactly represent data as low-dimensional embeddings that can be used as effective initialization for several underlying downstream tasks.

\paragraph{Contrastive Learning.}

The contrastive learning paradigm is popularly used to learn representations by comparing different samples in the dataset using distance metrics for structuring the latent space into similar and dissimilar embeddings \cite{mutinfo,cpc,initcontr}. \citet{simclr} utilized this idea to maximize the similarity between two views of the same input besides minimizing the similarity between the representations obtained from other images in a batch, leading to a stronger form of self-supervision. 

Contrastive learning aims at learning meaningful representations based on  positive and negative pairs of embeddings. Hence, in the case of iterative image retrieval we can model the positive and negative pairs through the selections made by the user, where all the pairs of similar images selected by a user can be modelled as positive pairs. On the other hand, the dissimilar images selected by the user act as the negative pair with all the similar images selected. Thus, using this as a weak supervision, we apply this concept to learn representations that map the notion of similarity specific to the user to a known similarity metric in the latent space.

\paragraph{Image Retrieval.}

The task of image retrieval using cues from users is a challenging task due to the high level of subjectivity and personalization in the user's conception of different visual features. \citet{g2g} used a higher form of supervision through natural language in order to retrieve images. Other approaches \cite{AttrMani,shapemani} relaxed these constraints by retrieving images correlated with the current query image based on a set of attributes that are either decided during system formulation or specified by the user. On similar lines, \citet{AttrMani} found the nearest set of orthogonal vectors as representatives for independent attributes in the latent space and weighted them by the attribute preferences to obtain modified query vectors. \citet{Bhattarai_2016_CVPR} explored the concept of learning similarity metrics in task-dependent projection spaces based on user feedback. Even though these systems vividly exploit some essential characteristics of facial images, they do not consider the information obtained from the previously chosen images selected by the users. 

In this work, we utilize the notion of similarity (or dissimilarity) as a form of user feedback, analogous to \cite{sgdfi,Bhattarai_2016_CVPR}. Through our framework, we highlight the significance of mapping the similarity notion of a particular user as distances in the latent space, enabling image recommendations closest to the user's mental image. \citet{Shrivastava2019FacefetchAE} adopted a similar approach for user personalization, however, it relied on labelled supervision.

\section{Proposed Approach}
\label{PropApp}

We propose a method in which the similarity and dissimilarity feedback can be viewed as positive and negative samples respectively to capture the notion of similarity in the user's mind. We attempt to associate this notion using a certain isotropic metric in the projected latent space of image representations. Thus, we put a constraint that the embeddings of images selected by the user as similar are closer than the ones selected as dissimilar in this low-dimensional space. We incorporate such a framework by projecting the pretrained base representations onto a lower-dimensional space using a fully connected neural network and formulate ${\tt SCLoss}$ to train the projection network in such a way that it learns to separate the projections of images relevant and irrelevant to the current query image. We also introduce the concept of anchoring during online training to conserve the notion of similarity and to ensure that only one cluster corresponding to the similar images is being formed and all dissimilar images are far from the cluster with a constant number of images required for training. 
We describe the our proposed relevance feedback algorithm ${\tt FaIRCoP}$ using Algorithm \ref{algo:RF} and highlight its essential components in the coming sections.

\begin{algorithm}
\caption{Relevance Feedback Algorithm}
\label{algo:RF}
\begin{algorithmic}
\STATE $embed \gets$ Encoder for extracting base embeddings for Images
\REQUIRE ${\tt Dossier}$, $k$, $u$, $prev\_samp$, $epochs$, $embed$
\STATE $S_{all}, D_{all} \gets \{\}$
\STATE $R \gets $Sample $k$ random images uniformly from sensitive attributes such as gender and complexion from ${\tt Dossier}$
\STATE $Rem \gets {\tt Dossier}$
\STATE $f \gets$ Initialize($f$)
\STATE $iter \gets 0$
\WHILE{Mental Image $\notin R$}
  \STATE $S \gets$ User Selects images from $R$
  \STATE $D \gets R - S$
  \STATE $Rem \gets Rem - R$
  \IF{$iter \% 2 ==0$}
      \STATE $S_{B} \gets sample(S_{all}, min(prev\_samp, |S_{all}|)) \cup S$
      \STATE $D_{B} \gets sample(D_{all}, min(prev\_samp, |D_{all}|)) \cup D$
      \STATE $i\gets0$
      \FOR{$i<epoch$}
      \STATE $S_{proj} \gets f(embed(S_B))$
      \STATE $D_{proj} \gets f(embed(D_B))$
      \STATE $L \gets ${\tt SCLoss}$(S_{proj}, D_{proj})$
      \STATE Backpropagate and Update the projection network $f$ to minimize $L$
      \STATE $i \gets i+1$
      \ENDFOR
  \ENDIF
  \STATE $S_{all} \gets S_{all} \cup S$
  \STATE $D_{all} \gets D_{all} \cup D$
  \FOR{$v \in Rem$}
    \STATE $sc_v \gets score(v, f(embed(S_{all})), f(embed(D_{all})))$
  \ENDFOR
  \STATE $R \gets$ Images corresponding to the top-$k$ scores
  \IF{$iter \% 3 ==0$}
    \STATE $R \gets R \cup sample(S_{all} \cup D_{all}, u)$
  \ELSE
    \STATE $R \gets R \cup sample(Rem, u)$
  \ENDIF
  \STATE $iter \gets iter+1$
\ENDWHILE
\end{algorithmic}
\end{algorithm}

\subsection{Disentangled Representation Learning}

Our relevance feedback framework requires good base representations to ensure that the encoder and the projection network do not have an additional overhead of jointly learning good representations, thus, leading to faster convergence and improved latency. Disentangled representations act as an effective initialization due to their ability to encapsulate disjoint factors of variation within specific fixed-sized chunks, resulting in enhanced downstream task performance.
Since we are dealing with a complex real-world dataset of images, it is not preferable to use labels to extract representations in the view of expensive annotations and limited expressivity. Hence, an unsupervised disentangled representation learning method is ideal for extracting representations for our task. We utilize \cite{Hu2018DisentanglingFO} (${\tt MIX}$) to extract representations for the database of images. We will illustrate the benefits of using disentangled representations for our framework in Section~\ref{sec:interpret}. 

\subsection{Separating Cluster Loss}

We propose a novel separating cluster loss (${\tt SCLoss}$) to create a cluster for the similar images specified by the user in the projected space and ensure that all the dissimilar images are farther from the similar images selected by the user. It is based on the notion of $N$-pair loss objective \cite{npairloss} which quantifies the loss for the objective of maximizing the similarity between a given pair of embeddings, known as a positive pair, and minimizing similarity with all the other embeddings. The loss equation for a positive pair $e$ and $e'$ along with a set $U$ consisting of all vectors, when paired up with e form negative pairs, with a scaling factor $\tau$\cite{simclr} is given by Equation~\ref{npairloss}.

\begin{equation}
\small
\label{npairloss}
    l_{U}(e, e') = - log \frac{e^{sim(e, e')/\tau}}{\sum_{k \in U} e^{sim(e, k)/\tau}}
\end{equation}

Hence, we can extend this notion into our setting, where, we maximize pairwise similarity between all the projected embeddings of similar images and ensure that all the projected embeddings of dissimilar images are farther from those of the similar images. It can be observed that our loss does not require an equal number of similar and dissimilar images and hence, can be flexibly used during online training.


\begin{equation}
\small
\label{loss}
    \begin{aligned}
        {\tt SCLoss}(S,D) = & \frac{1}{|S|(|S|-1)}\sum_{x \in S} \sum_{y \in S - \{x\}} l_{D}(x, y) \\
    \end{aligned}
\end{equation}

\subsection{Online Training and Inference}

The objective of the ${\tt SCLoss}$ is to train the projection network in such a way that the images which are similar to the mental image of the user would be nearer to the projected similar images cluster. Hence, we use a scoring function that has been specified in Equation~\ref{score}.

\begin{equation}
\small
\label{score}
    score(u) =  sim\Bigg(u,\frac{1}{|S_{a}|}\sum_{x \in S_{a}}x\Bigg) 
\end{equation}

It can be observed that as the number of iterations increase, there will be an increase in either the number of similar images, dissimilar images or both. In such cases, the computation required for the loss would become higher and may be computationally infeasible after a certain limit. Conversely, suppose only the similar and dissimilar images of the current iteration are used to create the clusters. In that case, clusters may be created for the two sets independent of similar previous images because there is no way to associate the new images with the previous selections. To circumvent this issue, we propose a training trick called \textit{anchoring} which ensure that the projected representations of new similar images are trained to be a part of the previously formed clusters in the projected space. During training, to ensure we have sufficient images to update our representations, we add a certain number of previously chosen similar and dissimilar images to the corresponding set obtained at the current iteration. This particular trick \textit{anchors} the new embeddings to the previously formed clusters as the network has already learnt to project the previous embeddings to the respective clusters avoiding the formation of multiple clusters in the latent space.


\section{Experiments}
We evaluate our approach against state-of-the-art baselines for the task of facial image retrieval using a set of qualitative and quantitative experiments.

\subsection{Dataset}

We utilize a set of two datasets, namely, Criminal Dataset and CelebA Dataset \cite{liu2015faceattributes}, to evaluate the efficacy of our approach against state-of-the-art in facial image retrieval approaches.

\subsubsection{Criminal Dataset:} The proposed framework was formulated with a goal to optimize the task of suspect image retrieval to assist criminal investigations based on the witness's mental image of the criminal. For the same, we utilize a criminal data dossier \cite{JainSeekSuspect} by the Metropolitan Police department of India containing $43,000$ unique criminal mugshots. Each data point in the dossier is associated with attributes describing the criminal’s physical attributes that include ${\tt face shape}$, ${\tt face complexion}$, etc. However, due to data privacy and confidentiality, we cannot release the dataset. 

\subsubsection{CelebA Dataset:} 
We employ the CelebA \cite{liu2015faceattributes} facial dataset in order to portray the generalizability and qualitative efficiency of our proposed framework.
The CelebA dataset contains $202,599$ facial images from $10,177$ identities. Each facial image is labeled with $40$ binary attributes, such as ${\tt pointy nose}$ and ${\tt wavy hair}$. However, we chose to discard all the recurring images of the same individual to maintain consistency with the Criminal Dataset which has a unique image for every individual. The CelebA dataset comprises of an exhaustive coverage of various ethnicities and genders, lending it wide acceptance among researchers.  

\subsection{Dataset Preprocessing} 
The criminal dataset had some noise in the form of alignment and blurriness. The disoriented images were correctly aligned using a pretrained VGGNet~\cite{conf/cvpr/SchroffKP15} finetuned on our dataset while the blurred images had to be discarded bringing the final dataset count to $39,196$ facial images. We then extracted the facial region from the mugshots using Haar Cascades \cite{conf/cvpr/ViolaJ01}. These preprocessing steps were not conducted on the CelebA dataset. 

\subsection{User Simulator}
We utilize a user simulator to make comparisons between different relevance feedback algorithms. The user simulator mimics a human user who has a target image in mind and provides feedback at each round. Each user simulation takes $30$ minutes to complete on average. We perform $10$ simulations for each algorithm and compare the average of the metrics over these simulations. We design a user simulator to replicate our user-in-the-loop framework.

\begin{algorithm}
\caption{User Simulator}
\label{algo:US}
\begin{algorithmic}
\REQUIRE ${\tt Dossier}$, ${\tt HOG}$, ${\tt FaceNet}$, ${\tt MIX}$, $w_{h}$, $w_{f}$,$w_{m}$, ${\tt RF\_Algo}$, ${\tt RF\_Algo\_Init}$
\STATE $Target \gets$ A random image sampled from ${\tt Dossier}$
\STATE $H_T, F_T, M_T  \gets$ ${\tt HOG}$, ${\tt FaceNet}$, ${\tt MIX}$ (Target)
\STATE $SampledImgs \gets sample(${\tt Dossier}$, min(|Dossier|, 1000)$
\STATE $thr \gets 0$
\FOR{$s \in SampledImgs$}
    \STATE $H_{s}, F_{s}, M_{s} \gets$ ${\tt HOG}$(s), ${\tt FaceNet}$(s), ${\tt MIX}$(s)
    \STATE $thr \gets thr + \frac{w_{h}sim(H_T, H_s) + w_{f}sim(F_T, F_s) + w_{m}sim(M_T, M_s)}{w_{h} + w_{f} + w_{m}}$
\ENDFOR
\STATE $thr \gets \frac{thr}{|SampledImgs|}$
\STATE $S \gets$ ${\tt RF\_Algo\_Init}$()
\STATE $iter, STemp \gets 0, \{\}$
\WHILE{Target $\notin S$}
  \STATE $Sim, DisSim \gets \{\}$
  \FOR{$s \in S$}
      \STATE $H_{s}, F_{s}, M_{s} \gets$ ${\tt HOG}$(s), ${\tt FaceNet}$(s), ${\tt MIX}$(s)
      \STATE $SimVal \gets \frac{w_{h}sim(H_T, H_s) + w_{f}sim(F_T, F_s) + w_{m}sim(M_T, M_s)}{w_{h} + w_{f} + w_{m}}$
      \IF{$SimVal > thr$}
        \STATE $Sim \gets Sim \cup \{s\}$
      \ELSE
        \STATE $DisSim \gets DisSim \cup \{s\}$
      \ENDIF
  \ENDFOR
  \STATE $STemp \gets STemp \cup Sim$
  \STATE $S \gets$ ${\tt RF\_Algo}$(Sim, DisSim)
  \STATE $iter \gets iter + 1$
  \IF{$iter \% 15 == 0$}
    \STATE $u \gets 0$
    \FOR{$s \in STemp$}
        \STATE $H_{s}, F_{s}, M_{s} \gets$ ${\tt HOG}$(s), ${\tt FaceNet}$(s), ${\tt MIX}$(s)
        \STATE $u \gets u + \frac{w_{h}sim(H_T, H_s) + w_{f}sim(F_T, F_s) + w_{m}sim(M_T, M_s)}{w_{h} + w_{f} + w_{m}}$
    \ENDFOR
    \STATE $u \gets \frac{u}{|STemp|}$
    \STATE $thr \gets 0.95thr + 0.05u$
    \STATE $STemp \gets \{\}$
  \ENDIF
\ENDWHILE
\end{algorithmic}
\end{algorithm}

To mimic the notion of similarity between two images, we design a metric comparing the mean of cosine similarities between the different representations of each of the images. If the similarity between the target image and the image under consideration is greater than a certain threshold, the simulator marks it as similar while the rest of the images are deemed dissimilar. We used a combinations of three image embeddings, namely, Histogram of Oriented Gradients (${\tt HOG}$) \cite{citeulike:335784}, ${\tt FaceNet}$ \cite{conf/cvpr/SchroffKP15}, and ${\tt MIX}$ \cite{Hu2018DisentanglingFO} to feed the representations in order to calculate the averaged similarity.

The threshold for similarity is determined at the start of each simulation by randomly sampling a constant number of images from the database and computing the similarity from the selected target image. These similarities are then averaged to get the initial threshold for similarity. After every constant number of iterations, the threshold is updated. The user simulator algorithm is detailed using Algorithm \ref{algo:US}.

\subsection{Metrics} \label{sssec:Metrics}

\paragraph{Percentile Ranking $\tt{(PR)}$.} Our use-case has a single unique solution to each query posed by different users. This solution is the target image which presents in a heavily filtered interval from a huge pool of data.
Hence, we calculate the ranking percentile of the target image, the magnitude of which is directly proportional to the accuracy of the model. 

\paragraph{Average Relevance $\tt{(AR)}$.}
We also use the Mean Average Relevance to quantify the relevance of the images suggested by the model in comparison to the user's mental image. It is indicative of how well the model is able to personalize according to the user's needs. We calculate the relevance for each simulation which is the fraction of similar images chosen by the user out of the total images displayed throughout the process. The mean of this over all the iterations gives us the average relevance.

\paragraph{Average Convergent Iterations $\tt{(ACI)}$.}
Due to the presence of a human-in-the-loop during relevance feedback, it is important to retrieve the target image in the lowest possible number of iterations. We calculate the average number of iterations it took to reach the target image for each simulation to obtain the average convergent iterations.

\subsection{Results on the User Simulator}
We compare the results on the simulator with FaceFetch \cite{Shrivastava2019FacefetchAE} and Rocchio Algorithm \cite{rocchio}, similar to an intermediate process mentioned in \cite{Shrivastava2019FacefetchAE} with the ${\tt MIX}$ embeddings. We calculate the metrics mentioned in Section \ref{sssec:Metrics} for all the aforementioned combinations of image embeddings. These results are displayed in Table \ref{tab:sim_criminal}.

\begin{figure}[h]
    \centering
      \begin{subfigure}{0.23\textwidth}
          \includegraphics[width=\textwidth]{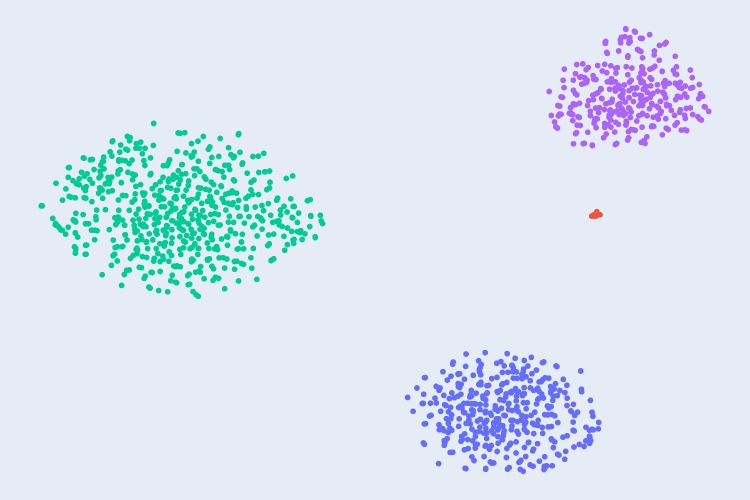}
          \caption{CelebA Dataset}
          \label{fig:celeba_user}
      \end{subfigure}
      \hfill
      \begin{subfigure}{0.23\textwidth}
          \includegraphics[width=\textwidth]{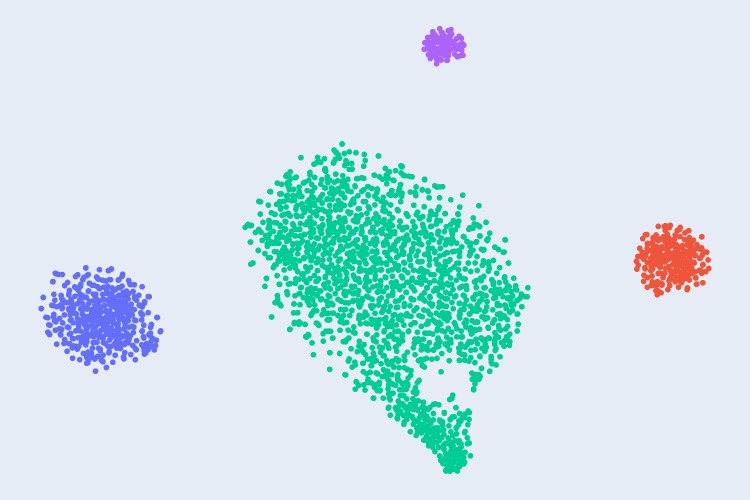}
          \caption{Criminal Dataset}
          \label{fig:criminal_user}
      \end{subfigure}
    \caption{Visualization for user-wise similarity preference clusters in the projected space using ${\tt FaIRCoP}$ for retrieving images on the simulator with $w_h = w_f = w_m = 1$. In these plots, each color depicts a distinct user.}
    \label{fig:userclusters}
\end{figure}

In Figure \ref{fig:userclusters}, we use t-SNE \cite{vanDerMaaten2008} to visualize the clusters of the similar images selected by the user simulator for four different simulations. Well defined clusters indicate that the notion of similarity for each user simulation was captured differently, thus, qualitatively expressing the level of personalization.

\begin{figure*}
  \centering
  \begin{subfigure}{0.24\textwidth}
      \includegraphics[width=\textwidth]{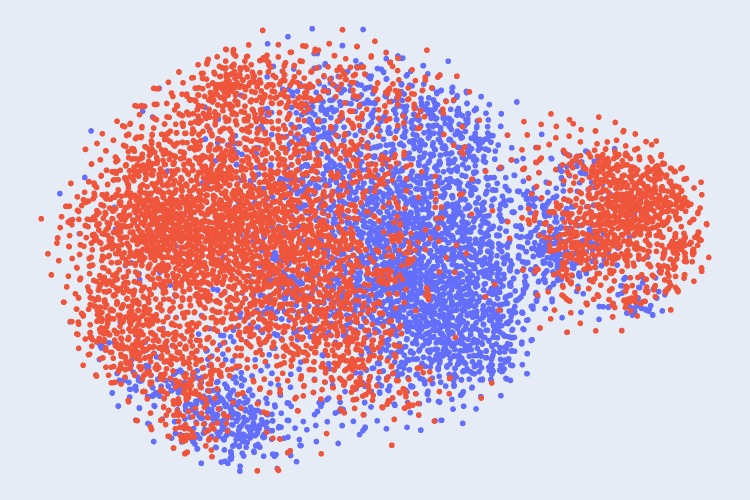}
      \caption{CelebA - Initial}
      \label{fig:celeba_wo_pretrain}
  \end{subfigure}
  \hfill
  \begin{subfigure}{0.24\textwidth}
      \includegraphics[width=\textwidth]{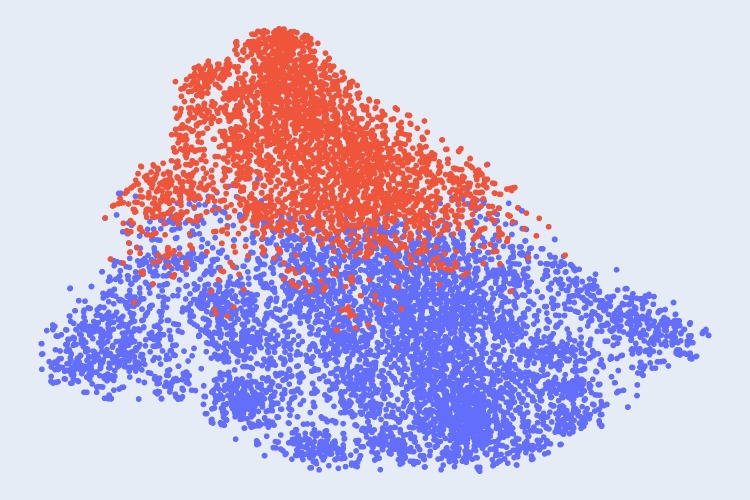}
      \caption{CelebA - Final}
      \label{fig:celeba_w_pretrain}
  \end{subfigure}
  \hfill
  \begin{subfigure}{0.24\textwidth}
      \includegraphics[width=\textwidth]{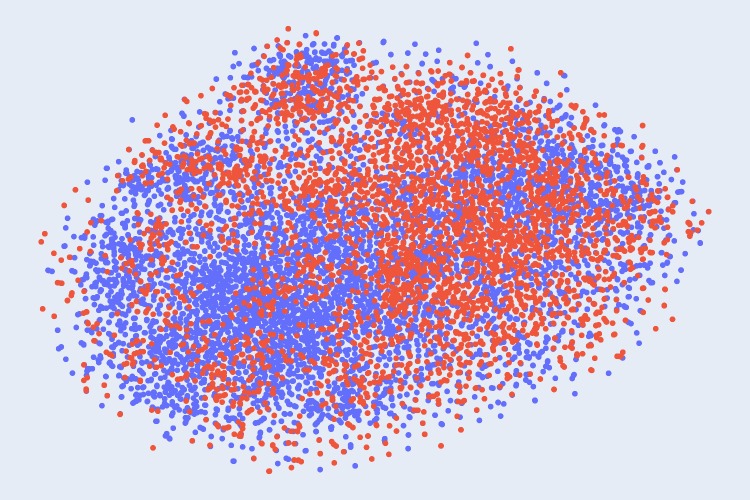}
      \caption{Criminal Data - Initial}
      \label{fig:celeba_wo_pretrain}
  \end{subfigure}
  \hfill
  \begin{subfigure}{0.24\textwidth}
      \includegraphics[width=\textwidth]{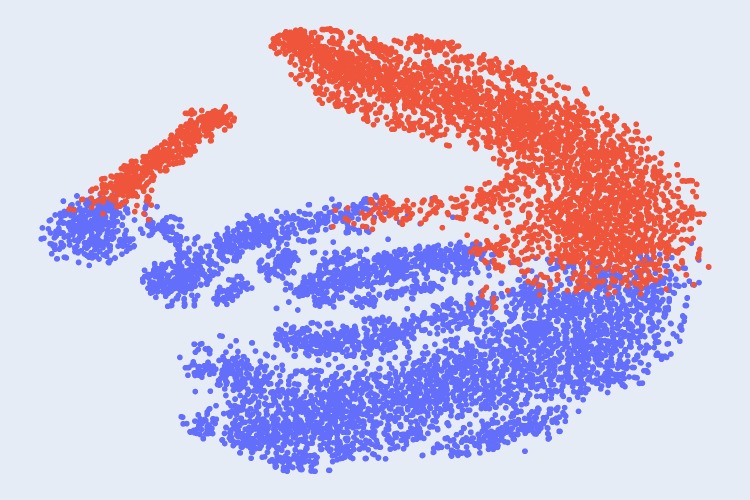}
      \caption{Criminal Data - Final}
      \label{fig:celeba_w_pretrain}
  \end{subfigure}
  \caption{
Visualization of projected embeddings of all similar  (blue) and dissimilar (red) images of the initial (top) and the trained (bottom) projection network for a simulation with ${\tt FaIRCoP}$ for both the datasets with $w_h = w_f = w_m = 1$.
}
\label{fig:simclusters}
\end{figure*}

Figure \ref{fig:simclusters} indicates that both the ${\tt FaIRCoP}$ is successful at contrasting between the similar and dissimilar images selected by the user simulator as the iterations proceed.

\begin{table}[b]
    \centering
    \begin{tabular}{|c|c|c|c|c|}
       \hline
         \textbf{Algorithm} & \textbf{${\tt PREF}$} & \textbf{${\tt REL}$} & \textbf{${\tt RESP}$} & \textbf{${\tt CONV}$}\\
      \hline
        \multicolumn{5}{|c|}{\textbf{Criminal Dataset}} \\ 
      \hline
        ${\tt Rocchio}$ & 0.40 & 0.41 & 0.41 & 0.08 \\
        ${\tt FaceFetch}$ & 0.40 & 0.59 & 0.57 & 0.28 \\
        ${\tt FaIRCoP}$ & \textbf{0.70} & \textbf{0.9} & \textbf{0.85} & \textbf{0.44} \\
      \hline
        \multicolumn{5}{|c|}{\textbf{CelebA Dataset}} \\ 
        \hline
        ${\tt Rocchio}$ & 0.26 & 0.43 & 0.42 & 0.14\\
        ${\tt FaceFetch}$ & 0.5 & 0.63 & 0.58 & 0.29\\
        ${\tt FaIRCoP}$ & \textbf{0.73} & \textbf{0.86} & \textbf{0.9} & \textbf{0.40}\\
        \hline
    \end{tabular}
    \caption{Cumulative metrics obtained from the User Study conducted on Criminal and CelebA dataset.}
    \label{tab:user_study}
\end{table}

\begin{table}[h]
    \centering
        \begin{tabular}{|c||c|c||c|c|}
            \hline
                \textbf{Metric} & \multicolumn{2}{|c||}{\textbf{Criminal Dataset}} & \multicolumn{2}{|c|}{\textbf{CelebA Dataset}} \\
                \hline
                 & \textbf{${\tt ResNet}$} & \textbf{${\tt FaIRCoP}$} & \textbf{${\tt ResNet}$} & ${\tt FaIRCoP}$ \\
            \hline
                $\mathcal{D}$ & 0.23 & \textbf{0.30} & 0.27 & \textbf{0.36} \\
                $\mathcal{C}$ & 0.12 & \textbf{0.15} & 0.21 &  \textbf{0.27}\\
                $\mathcal{I}$ & \textbf{0.89} & \textbf{0.89} & 0.88 & \textbf{0.90} \\
            \hline
        \end{tabular}
    \caption{Interpretability metric score comparison for ${\tt ResNet}$ and ${\tt FaIRCoP}$ embedddings.}
    \label{tab:dci}
\end{table}

\begin{table*}[h]
    \centering
    \renewcommand{\tabcolsep}{2pt}
        \begin{tabular}{|c|c|c||c|c|c||c|c|c||c|c|c|}
            \hline
            
                \multicolumn{3}{|c||}{\textbf{Representation}} & \multicolumn{3}{c||}{\textbf{ $\tt{ACI}$}} & \multicolumn{3}{c||}{\textbf{${\tt AR}$}} & \multicolumn{3}{c|}{\textbf{ ${\tt PR}$}} \\  
                \hline
                
                \multicolumn{12}{|c|}{\textbf{Criminal Dataset}} \\
                
                \hline
            
                \textbf{${\tt FaceNet}$} & {${\tt MIX}$} & \textbf{${\tt HOG}$} & \textbf{${\tt Rocchio}$} & \textbf{${\tt FaceFetch}$} & \textbf{${\tt FaIRCoP}$}& 
                                                                                  \textbf{${\tt Rocchio}$} & \textbf{${\tt FaceFetch}$} & \textbf{${\tt FaIRCoP}$}& 
                                                                                  \textbf{${\tt Rocchio}$} & \textbf{${\tt FaceFetch}$} & \textbf{${\tt FaIRCoP}$}\\
            \hline
              \checkmark & \checkmark & \checkmark & 804.22 & 691.00 & \textbf{57.25} & 0.29 & 0.15 & \textbf{0.82} & 0.71 & 0.77 & \textbf{0.98}  \\
              \checkmark & \checkmark & & 450.80 & 506.50  & \textbf{68.33} & 0.45 & 0.27 & \textbf{0.83} & 0.83 & 0.82 & \textbf{0.99} \\
              & \checkmark & \checkmark & 550.95 & 152.50  & \textbf{41.66}  & 0.52 & 0.38 & \textbf{0.79} & 0.80 & 0.95 & \textbf{0.99} \\
              \checkmark & & \checkmark &  565.60 & 457.75  & \textbf{98.33} & 0.34 & 0.17 & \textbf{0.79} & 0.80 & 0.72 & \textbf{0.98}\\
              & \checkmark & & 441.30 & 380.75  & \textbf{89.00} & 0.59 & 0.36 & \textbf{0.88} & 0.84 & 0.86 & \textbf{0.96} \\
              
            \hline
            
            \multicolumn{12}{|c|}{\textbf{CelebA Dataset}} \\
                
                \hline
                \textbf{${\tt FaceNet}$} & {${\tt MIX}$} & \textbf{${\tt HOG}$} & \textbf{${\tt Rocchio}$} & \textbf{${\tt FaceFetch}$} & \textbf{${\tt FaIRCoP}$}& 
                                                                                  \textbf{${\tt Rocchio}$} & \textbf{${\tt FaceFetch}$} & \textbf{${\tt FaIRCoP}$}& 
                                                                                  \textbf{${\tt Rocchio}$} & \textbf{${\tt FaceFetch}$} & \textbf{${\tt FaIRCoP}$}\\
            \hline
              \checkmark & \checkmark & \checkmark & 351.2 & 263.00 & \textbf{40.5}& 0.37 & 0.40 & \textbf{0.61} & 0.44  & 0.58 & \textbf{0.97}\\
              \checkmark & \checkmark & & 358.8 & 222.8 & \textbf{27.4}& 0.36 & 0.40 & \textbf{0.70} & 0.43 & 0.65 & \textbf{0.96}\\
               & \checkmark & \checkmark & 299.8 & 255.20 & \textbf{50.0}&  0.54 & 0.37 & \textbf{0.87} & 0.52 & 0.59 & \textbf{0.92} \\
              \checkmark &  & \checkmark & 309.00 & 249.00 & \textbf{98.2} & 0.36 & 0.38 & \textbf{0.54} & 0.51 & 0.60 & \textbf{0.84} \\
               & \checkmark &  & 158.4 & 227.00 & \textbf{20.2} & 0.53 & 0.43 & \textbf{0.82} & 0.75 & 0.64 & \textbf{0.97}  \\
            \hline
            
        \end{tabular}
    \caption{Quantitative metrics obtained from user simulation using different methods on the Criminal and CelebA dataset.}
    \label{tab:sim_criminal}
\end{table*}

\section{User Study}
We conducted a user study to test the efficacy of our method in real-life scenarios and compare the performance with baseline methods for iterative image retrieval using relevance feedback. The study involved $20$ participants, with each of them assigned an image from the database. Each user was displayed their image for $40$ seconds to generate a suitable visual memory of the image assigned. Based on their visual memory, they searched the image using four separate systems with the ${\tt FaIRCoP}$, ${\tt Rocchio}$ \cite{rocchio}, and ${\tt FaceFetch}$ \cite{Shrivastava2019FacefetchAE} running at the respective backends. For each search, the users were initially required to select the attributes used to initialize the search with a random set of images that had suitable similarities to the attributes provided. The algorithms showed users $16$ images at each iteration, from which they selected similar images and got a recommendation for the new images through the methods mentioned above. This process was repeated until the user reported an image that matched heavily with the user's visual memory. The searches were clipped to a maximum of $30$ iterations in case of the criminal dataset \cite{JainSeekSuspect}. In contrast, they were clipped for a maximum of $25$ iterations in the case of CelebA \cite{liu2015faceattributes}. At the end of each search, the users were asked to fill a questionnaire based on which some performance measures were computed as mentioned in sub-section \ref{ssec:perf_meas}. The averaged results for all users are depicted in Table \ref{tab:user_study}. The user study results correlated significantly with the simulation results we obtained regarding the metrics employed.

\begin{figure}
    \centering
    \includegraphics[scale=0.4]{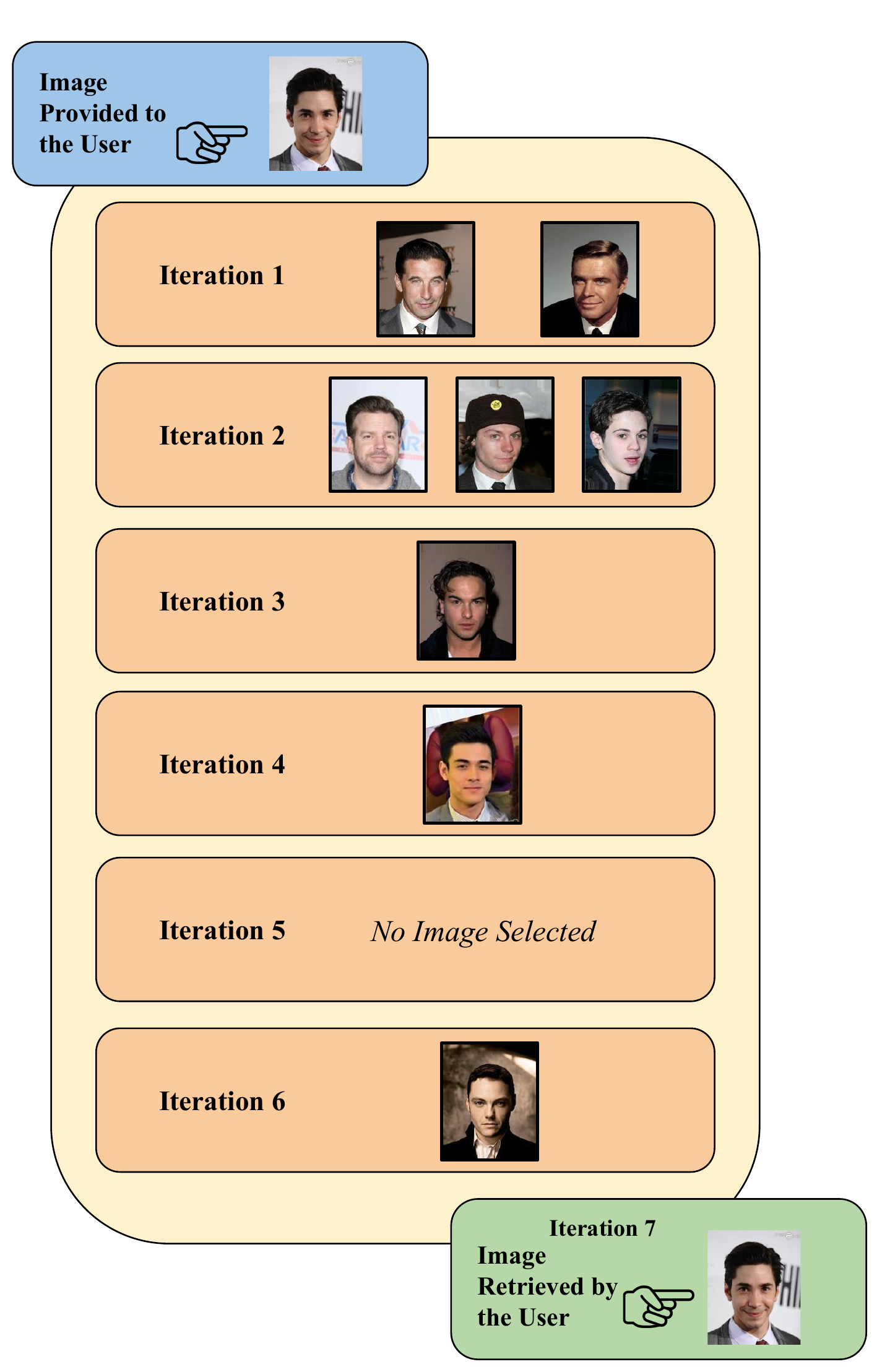}
    \caption{Another run obatined from the user study using ${\tt FaIRCoP}$. Each box contains similar images selected by the user at each iteration from recommended images until convergence.}
    \label{fig:faircop_np}
\end{figure}

\subsection{Performance Metrics} \label{ssec:perf_meas}
Due to the presence of a human-in-the-loop during the relevance feedback mechanism, we conduct a user study and evaluate the performance of our model compared to other baselines based on the post-study questionnaire, which covers the user feedback on the metrics discussed below.

\paragraph{Relevance (${\tt REL}$).} Due to the iterative nature of the mechanism, it is essential that there must be an increasing similarity between the user's visual image and the images recommended as the iterations proceed. The relevance of a system measures the change in similarity perceived by the user between the set of images recommended at each iteration and the visual memory of the user. We asked the users to quantify their ease of selecting similar images as iterations proceeded on a scale of $1$ to $5$, where $1$ denoted high-level mental stress in selection whereas $5$ denoted increasing ease as iterations proceeded. We normalized the score to lie between $0$ and $1$.

\paragraph{Responsiveness (${\tt RESP}$).} For iterative image retrieval, it is also essential that the users observe that their previous responses are being used effectively and the images are not randomly recommended. We asked the users to quantify their perceived randomness of recommendations on a scale of $1$ to $5$, where $1$ denoted a high amount of randomness whereas $5$ denoted an effective use of previous queries. We normalized the score to lie between $0$ and $1$.

\paragraph{Convergence (${\tt CONV}$).} As a system for image retrieval, we must ensure that the system can converge in fewer iterations. To measure this, we calculate convergence ($C$) for a search converged in $N$ iterations with a maximum limit of ${\tt max\_iter}$ allowed as given in Equation~\ref{converge}.

\begin{equation}
   C = 
        \begin{cases}
            1 - \frac{N}{{\tt max\_iter} + 5} & \text{if user reports image}\\
            0              & \text{otherwise}
        \end{cases}
   \label{converge}
\end{equation}

\paragraph{Preferability (${\tt PREF}$).} Since each user performed the image retrieval on all the algorithms, we asked them to report their willingness to continue using the system if the retrieved images did not match the exact target image. To capture the user's preferablity to use the system. We assigned a preferability score of $0$ if the user was unwilling to continue, $0.75$ if the user wanted to continue, and $1$ if the user retrieved the exact image.

\section{Additional Studies}
We mention some additional studies that we performed which provide the process which was involved behind obtaining the {\tt$SCLoss$} and the formulation of the algorithm. Moreover, these studies also give an insight on certain alternatives that could have been used for this algorithm but contained caveats which were revealed during experiments and the user study.

\subsection{Pretraining}
The parameters of the projection network can also be initialized randomly at the start of the search or obtained through pretraining the projection network. The projection network can be pretrained in the same way as the training is done to learn representations in \citet{simclr} with the parameters of the encoder being frozen and the projection network being highly regularized. The pretraining projection network did not show much improvement in the performance on both the simulation and the user study as compared to a projection network which was randomly initialized at the start of the search. We show experiments comparing both the pretrained (${\tt FaIRCoP-P}$) and the non-pretrained (${\tt FaIRCoP}$) variants of our algorithm in the supplementary.

\subsection{Alternate Loss Function}
An alternative of the {\tt$SCLoss$} and the scoring function introduced in Section~\ref{PropApp} can be written as Equation~\ref{altloss} and Equation~\ref{scorealt} respectively.

\begin{equation}
\small
\label{altloss}
    \begin{aligned}
        L_{s}(S,D) = & \sum_{x \in S} \sum_{y \in S - \{x\}} l_{D}(x, y) \\
        L_{d}(S,D) = & \sum_{x \in D} \sum_{y \in D - \{x\}} l_{S}(x, y) \\
        {\tt SCLoss}_{alt}(S, D) = & \frac{L_{s}(S,D)}{2|S|(|S|-1)} + \frac{L_{d}(S,D)}{2|D|(|D|-1)}
    \end{aligned}
\end{equation}

In this function, the definition of $l$ remains the same as defined originally in Section~\ref{PropApp}. This function establishes an explicit dichotomy between the similar images and dissimilar images and forces the projection network to cluster them separately as compared to the original loss function which relaxed the clustering constraint on the dissimilar set of images. Consequently, this change would accordingly modify the scoring function which would also be characterized by its distance from the cluster containing the projected embeddings of dissimilar images selected by the user as given in Equation~\ref{scorealt}.

\begin{equation}
\small
\label{scorealt}
    score_{alt}(u) =  sim(u,\frac{1}{|S_{a}|}\sum_{x \in S_{a}}x) -  sim(u, \frac{1}{|D_{a}|}\sum_{x \in D_{a}}x)
\end{equation}

This loss function gave very similar results on the simulations but proved to be inefficient when it was employed for the user study. We speculated that this problem arises due to the practical problem of encountering dissimilar images of a higher number in the initial iterations as compared to similar images. Thus, constraining the dissimilar images to a single cluster leads to learning degenerate solutions where the latent space is mostly covered by the dissimilar images, hence, not leading to the desired separation between both the clusters. Moreover, our problem hypothesizes that all the similar images have certain common features which cause the users to select an image as similar to their mental image, but with clustering all the dissimilar images, we essentially force the projection network to project large feature differences nearer to each other, which may cause the projection network to learn less meaningful projections, thus, leading to delayed convergence in real time. With relaxing the dissimilarity constraint, we obtained better results with the user study.

\section{Interpretability and Fairness}
\label{sec:interpret}

\begin{figure}
    \centering
    \includegraphics[scale=0.22]{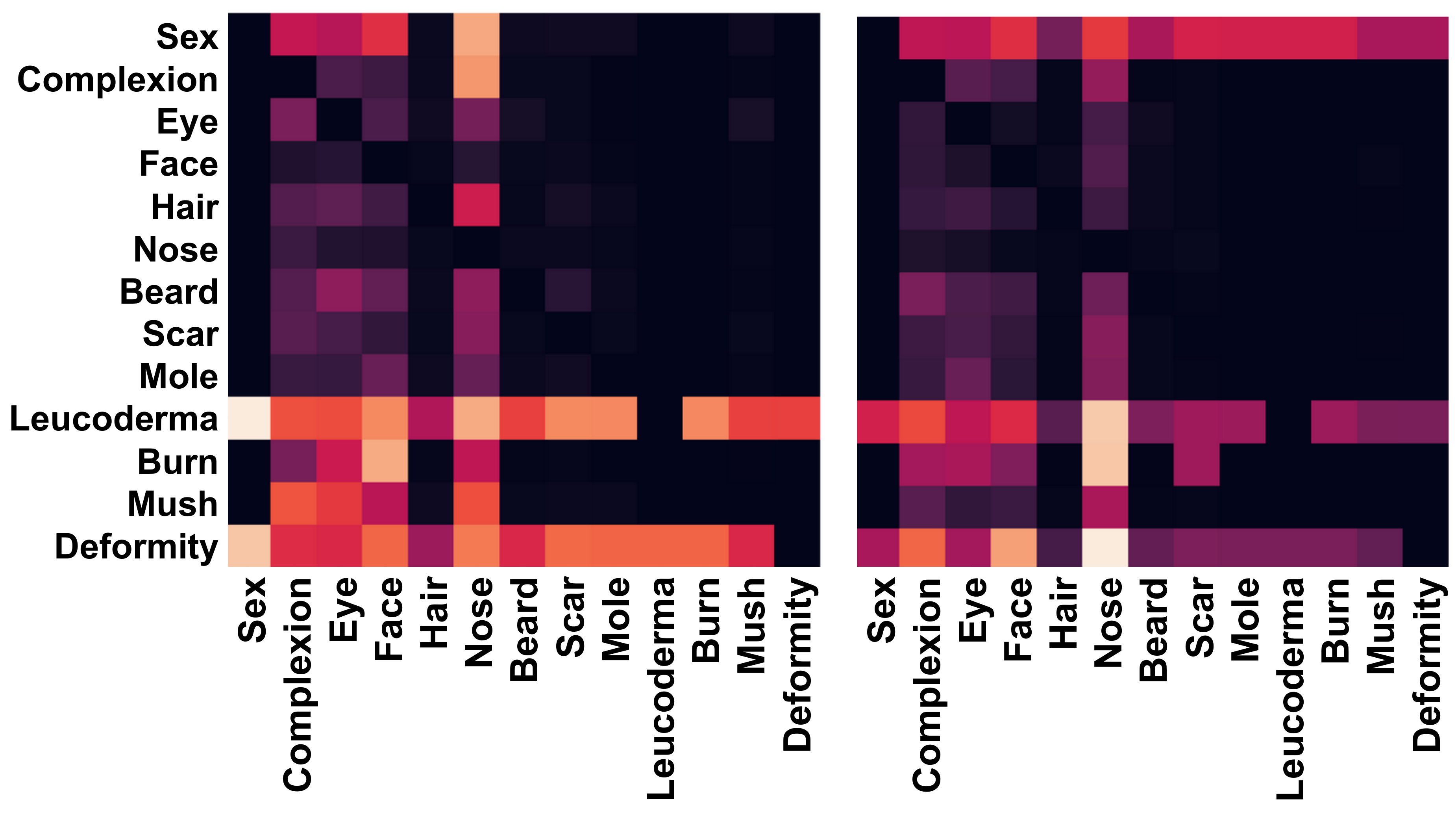}
    \caption{Modified Group Demographic Parity metric for comparing the fairness of ${\tt FaIRCoP}$ (left, $\mathcal{F} = 0.04$) and ${\tt ResNet}$ (right, $\mathcal{F} = 0.05$).}
    \label{fig:dip}
\end{figure}

For any given dataset, semantically meaningful feature learning requires the learnt representations to be interpretable. Due to the interactive nature of the problem, the learnt representations should be fair to avoid introducing any feedback bias among users interacting with the system.

\subsection{Interpretability}
We evaluated the interpretability of the representations for all the datasets using the Disentanglement ($\mathcal{D}$), Completeness ($\mathcal{C}$), and Informativeness ($\mathcal{I}$) (DCI) metric \cite{DCI}. A high value for each of these metrics depicts a high semantic meaning correlated with the tangible features in the dataset\cite{DCI_1, DCI_2}. Considering $\mathbb{F}$ as the total number of factors of variation in the dataset, we trained $\mathbb{F}$ gradient boosting regressors for each representation in the dataset as the feature set and generated an importance matrix $R$ such that for a given latent factor $j$, $R_{i,j}$ represents the $i$th feature importance weight of the linear regressor trained on the set of representations with $j$th factor of variation in the output.

\paragraph{Disentanglement ($\mathcal{D}$).}
The disentanglement score of the metric represents the degree to which a given representation disentangles the underlying factors of variation. The total disentanglement score for the factor of variations was calculated as follows:
\begin{equation}
    \mathcal{D} = \sum^{}_{i} \Big(1 - H(P_i)\Big) \frac{\sum^{}_{j}R_{i,j}}{\sum^{}_{i,j}R_{i,j}},
\end{equation}

where $H(P_i)$ represents the entropy of the $P_i$ distribution where, $P_i$ is a $j \times 1$ vector such that $P_{i,j} = \frac{R_{i,j}}{\sum^{}_{k}R_{i,k}}$. The score directly represents disentanglement as it is equal to $1$ only when each representation is deemed important for predicting only $1$ out of the different factors of variations.

\paragraph{Completeness ($\mathcal{C}$).}
The completeness score measures the degree to which a single factor of variation $j$ is captured by the representations and is calculated as follows:
\begin{equation}
    \mathcal{C}_j = 1 - H(P_j),
\end{equation}
where $H(P_j)$ is calculated in the same manner as described in the previous section. The score $C_j$ is equal to $1$ if only one representation is important for predicting the $j$th factor of variation and is equal to $0$ if all representations contribute equally. The final score is calculated as follows for all the $\mathbb{F}$ factors of variations:
\begin{equation}
    \mathcal{C} = \sum^{\mathbb{F}}_{j=1} C_j.
\end{equation}

\paragraph{Informativeness ($\mathcal{I}$).}
The information score represents the degree of information captured by a representation for all the underlying factors of variations and is calculated as follows:
\begin{equation}
    \mathcal{I} = \mathbb{E}_{j \in (\mathbb{Z} \cap [1\ldots\mathbb{F}])} \Big[1 - \| z_j - z_j'\|\Big],
\end{equation}
where $z_j$ represents the true distribution of the $j^{th}$ factor of variation and $z_j'$ represents the distribution predicted by the $j^{th}$ linear regressor. Ignoring the dependence of this metric on the capacity of the used regressors, this metric is equal to 1 when the representations are perfectly able to predict all the factors of variations.

Table \ref{tab:dci} illustrates that the ${\tt FaIRCoP}$ embeddings outperform the embeddings of the pretrained ${\tt ResNet}-18$ model across all three metrics for both the datasets.

\subsection{Fairness} 
 Since we do not release the criminal dataset to maintain confidentiality, we illustrate an extensive fairness study on the dataset to provide an idea about the label distribution in the dataset. To evaluate the efficacy of the representation generators in terms of fairness, we evaluated them based on two metrics -- \textit{group fairness} \cite{fairness} and \textit{label distribution similarity}. We used a modified group demographic parity metric \cite{fairness} and depict the results that we obtained in Figure~\ref{fig:dip}. For group fairness, we used a custom demographic parity based measure, where, a low metric value (which is essentially an average of the pairwise differences for each pair in the sample set spanned by the joint distribution of $t$ and $s$) indicates equal characterization for each minority group (represented by $s$) with respect to each target group (represented by $t$). The steps involve dividing the generated representations into training and testing sets and generating each possible pair of factor of variations as sensitive variable $s$ and target variable $t$ consecutively. For each pair, a $K$-Nearest Neighbors (KNN) clustering model was fit onto the training set with $t$ as the output and the conditional probability $p(t_i|s_j)$ for all $t_i \in t$ and $s_j \in s$ was calculated to get the final heatmap. To summarize the heatmap in a single quantitative metric, we evaluated the average value using the following formulation.

\begin{equation}
    \mathcal{F} = \mathbb{E}_{[t_i, s_j] \in [t,s]} \big[p(t_i|s_j)\big].
\end{equation}

The results can be found in Fig. \ref{fig:dip}, where, we evaluated FaIRCoP embeddings against a pretrained ${\tt ResNet}-18$ on our entire training set.

We also evaluated the distribution similarity between the dissimilar images selected by the simulator with the entire training set to evaluated any biases in the clustering procedure. The results for distribution similarity can be found in the supplementary material.

\section{Conclusion}
In this work, we tackled the problem of facial image retrieval using contrastive learning over user feedback which serves as a weaker form of supervision for the system. For this, we propose the {\tt$SCLoss$}, along with a pretraining and online inference strategy. Our system caters to the personalized notion of features that each user has due to high subjectivity in the mental visual memory of a person. Equipped with a user-friendly web interface, our proposed algorithm outperforms other state-of-the-art baselines qualitatively and quantitatively, as verified through the user study. 

\bibliographystyle{ACM-Reference-Format}
\bibliography{main}

\end{document}


\title{FaIRCoP: Facial Image Retrieval using Contrastive Personalization \\ (Supplementary Material)}


\author{Anonymous}

\renewcommand{\shortauthors}{Anonymous}

\maketitle

\begin{table}[t]
    \centering
    \begin{tabular}{|c|c|c|c|c|}
       \hline
         \textbf{Algorithm} & \textbf{${\tt PREF}$} & \textbf{${\tt REL}$} & \textbf{${\tt RESP}$} & \textbf{${\tt CONV}$}\\
      \hline
        \multicolumn{5}{|c|}{\textbf{Criminal Dataset}} \\ 
      \hline
        ${\tt FaIRCoP}$ & 0.70 & 0.9 & 0.85 & 0.44 \\
        ${\tt FaIRCoP - P}$ & 0.68 & 0.91 & 0.82 & 0.45 \\
      \hline
        \multicolumn{5}{|c|}{\textbf{CelebA Dataset}} \\ 
        \hline
        ${\tt FaIRCoP}$ & 0.73 & 0.86 & 0.9 & 0.37 \\
        ${\tt FaIRCoP - P}$ & 0.75 & 0.87 & 0.91 & 0.43 \\
        \hline
    \end{tabular}
    \caption{Cumulative metrics obtained from the User Study conducted on Criminal and CelebA dataset.}
    \label{tab:user_study}
\end{table}

\begin{table*}[b]
    \footnotesize
    \centering
    \renewcommand{\tabcolsep}{2pt}
        \begin{tabular}{|c|c|c||c|c||c|c||c|c|}
            \hline
            
                \multicolumn{3}{|c||}{\textbf{Representation}} & \multicolumn{2}{c||}{\textbf{ $\tt{ACI}$}} & \multicolumn{2}{c||}{\textbf{${\tt AR}$}} & \multicolumn{2}{c|}{\textbf{ ${\tt PR}$}} \\  
                \hline
                
                \multicolumn{9}{|c|}{\textbf{Criminal Dataset}} \\
                
                \hline
            
                \textbf{${\tt FaceNet}$} & {${\tt MIX}$} & \textbf{${\tt HOG}$} & \textbf{${\tt FaIRCoP}$} &\textbf{${\tt FaIRCoP-P}$} & 
                                                                     \textbf{${\tt FaIRCoP}$} & \textbf{${\tt FaIRCoP-P}$} & 
                                                                          \textbf{${\tt FaIRCoP}$} & \textbf{${\tt FaIRCoP-P}$} \\
            \hline
              \checkmark & \checkmark & \checkmark & \textbf{57.25} & 87.20 & 0.82 & \textbf{0.83} & \textbf{0.98} & 0.97  \\
              \checkmark & \checkmark & & \textbf{68.33} & 73.10 & \textbf{0.83} & 0.82 & \textbf{0.99} & 0.97 \\
              & \checkmark & \checkmark & \textbf{41.66} & 74.90 & 0.79 & \textbf{0.88} & \textbf{0.99} & 0.97 \\
              \checkmark & & \checkmark & \textbf{98.33} & 100.75 & \textbf{0.79} & 0.64 & \textbf{0.98} & 0.86 \\
              & \checkmark & & 89.00 & \textbf{74.40} & \textbf{0.88} & \textbf{0.88} & 0.96 & \textbf{0.97}  \\
              
            \hline
            
            \multicolumn{9}{|c|}{\textbf{CelebA Dataset}} \\
                
                \hline
            \textbf{${\tt FaceNet}$} & {${\tt MIX}$} & \textbf{${\tt HOG}$} & \textbf{${\tt FaIRCoP}$} &\textbf{${\tt FaIRCoP-P}$} & 
                                                                         \textbf{${\tt FaIRCoP}$} & \textbf{${\tt FaIRCoP-P}$} & 
                                                                         
                                     \textbf{${\tt FaIRCoP}$} & \textbf{${\tt FaIRCoP-P}$} \\
            \hline
              \checkmark & \checkmark & \checkmark & 40.5 & \textbf{37.4} & 0.61 & \textbf{0.68} & \textbf{0.97} & 0.94 \\
              \checkmark & \checkmark & & \textbf{27.4} & 76.6 & \textbf{0.70} & 0.61 & \textbf{0.96} & 0.88 \\
               & \checkmark & \checkmark & \textbf{50.0} & 58.0 & \textbf{0.87} & 0.83 & \textbf{0.92} & 0.91 \\
              \checkmark &  & \checkmark & 98.2 & \textbf{62.6} & 0.54 & \textbf{0.55} & 0.84 & \textbf{0.90} \\
               & \checkmark &  & \textbf{20.2} & 44.0 & 0.82 & \textbf{0.87} & \textbf{0.97} & 0.93 \\
            \hline
            
        \end{tabular}
    \caption{Quantitative metrics obtained from user simulation using different methods on the Criminal and CelebA dataset.}
    \label{tab:sim_criminal}
\end{table*}

\section{Effect of Pretraining}
As discussed in the paper, we present the user study results and simulation results comparing the pretrained and non-pretrained versions of FaIRCoP as given in Table~\ref{tab:user_study} and Table~\ref{tab:sim_criminal}. We can see that there are no signigicant differences in both the methods.

\section{Distribution Similarity Interpretation}

\begin{figure*}
    \centering

    \subcaptionbox{Beard}{
    \begin{subfigure}[t]{.35\linewidth}
    \centering
     \includegraphics[width=1.0\textwidth]{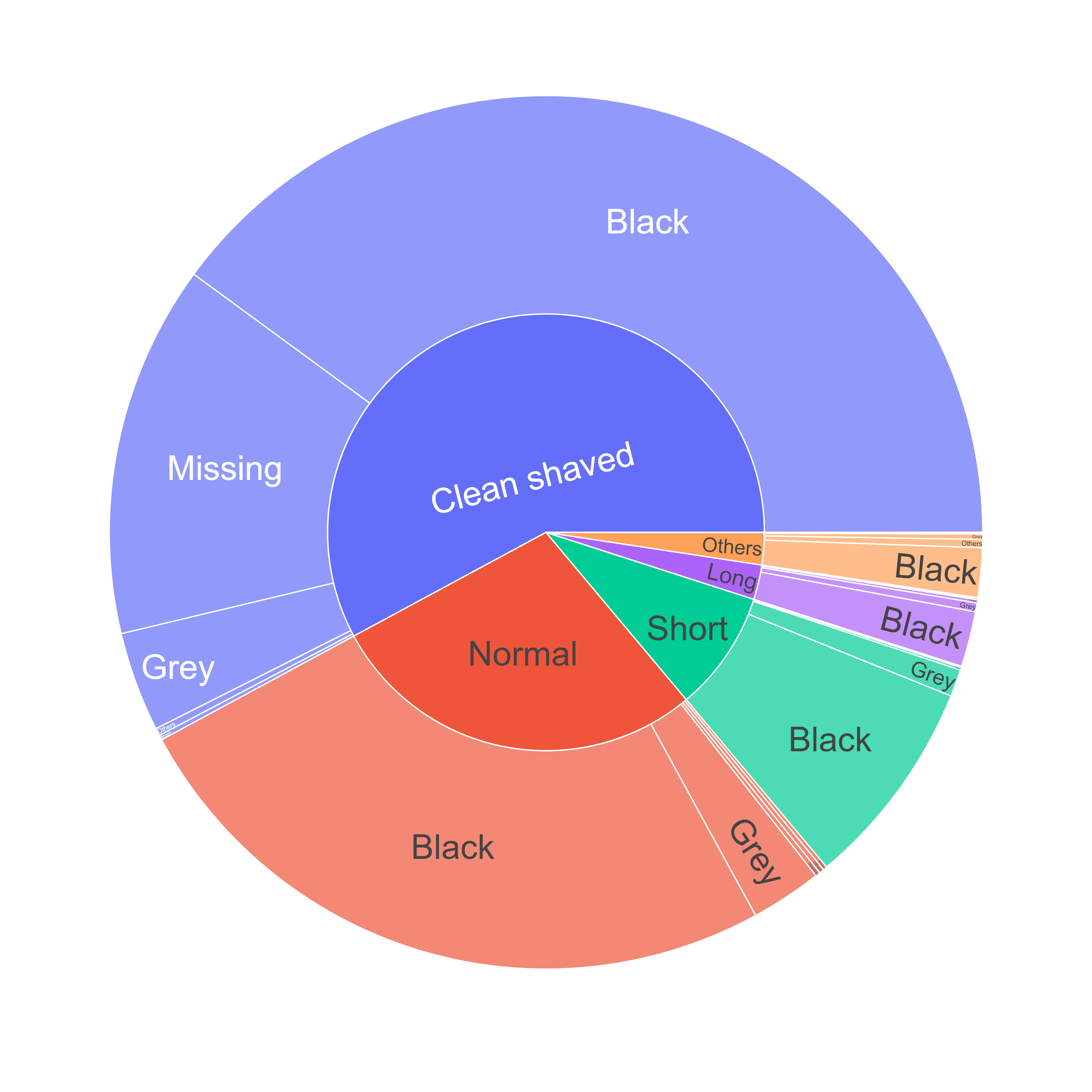}
     \caption*{Full Dataset}
  \end{subfigure}
  \begin{subfigure}[t]{.35\linewidth}
    \centering
     \includegraphics[width=1.0\textwidth]{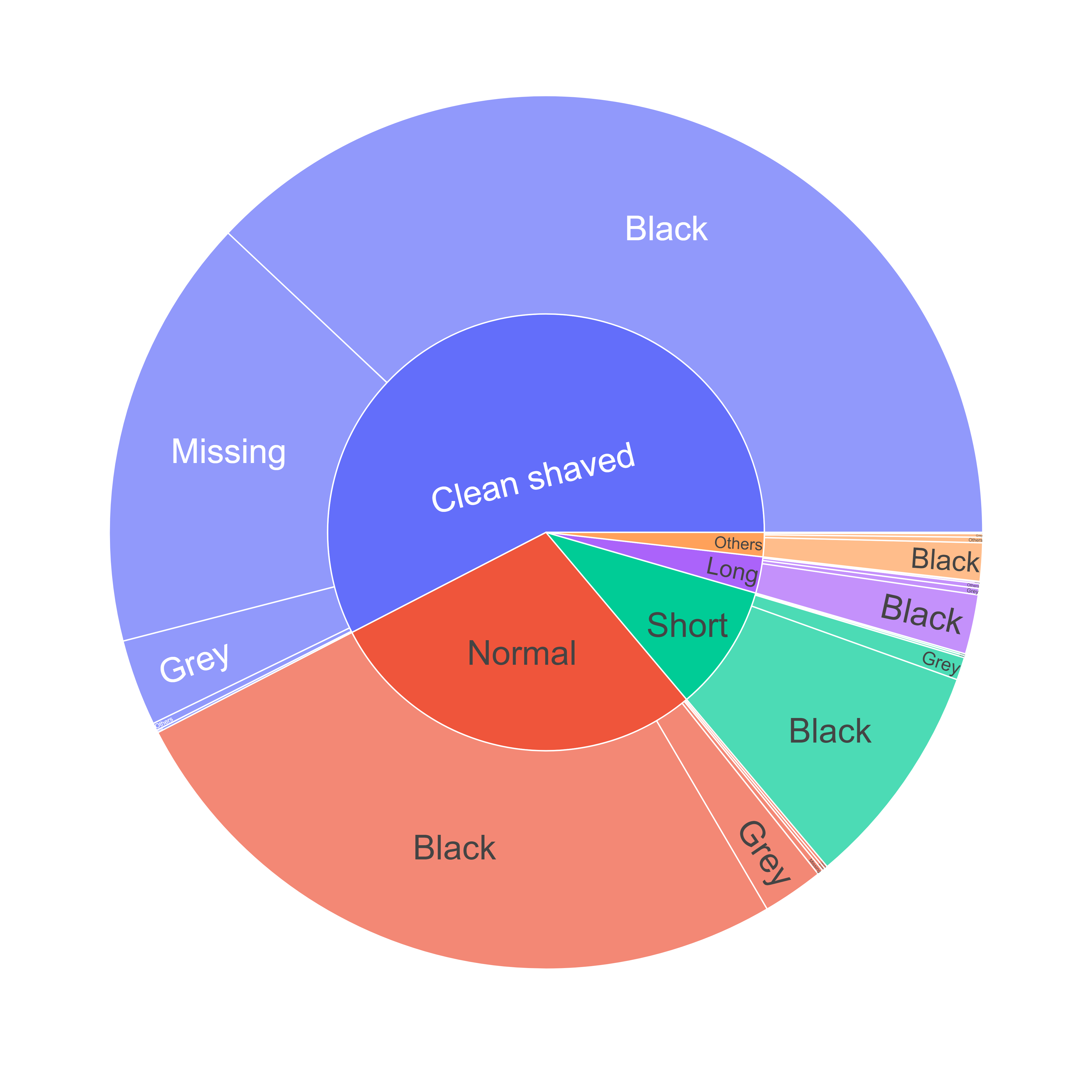}
     \caption*{Selected Dissimilar Images}
  \end{subfigure}
  }
  
  \subcaptionbox{Facial Characteristics}{
    \begin{subfigure}[t]{.35\linewidth}
    \centering
     \includegraphics[width=1.0\textwidth]{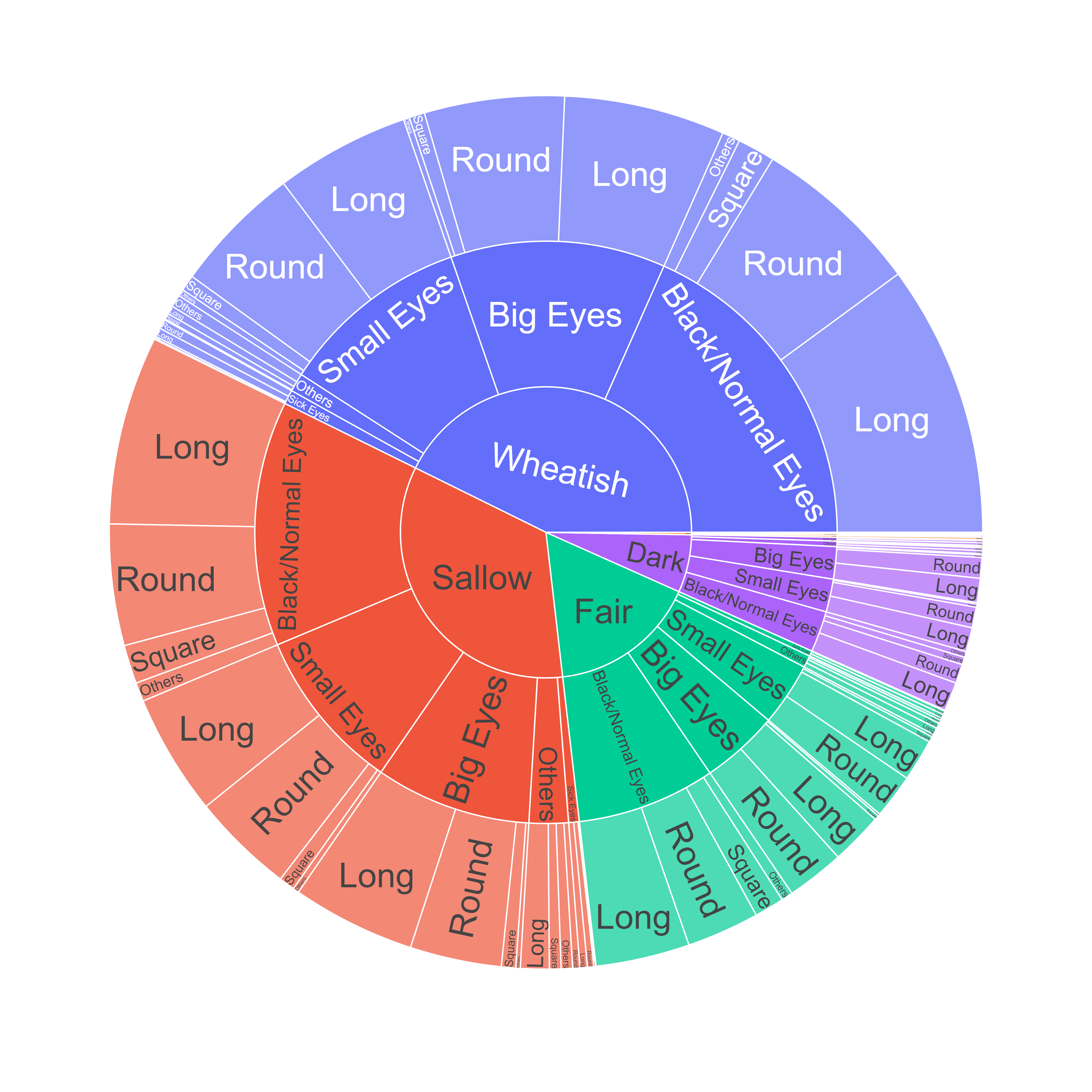}
    \caption*{Full Dataset}
    \end{subfigure}
    \begin{subfigure}[t]{.35\linewidth}
    \centering
     \includegraphics[width=1.0\textwidth]{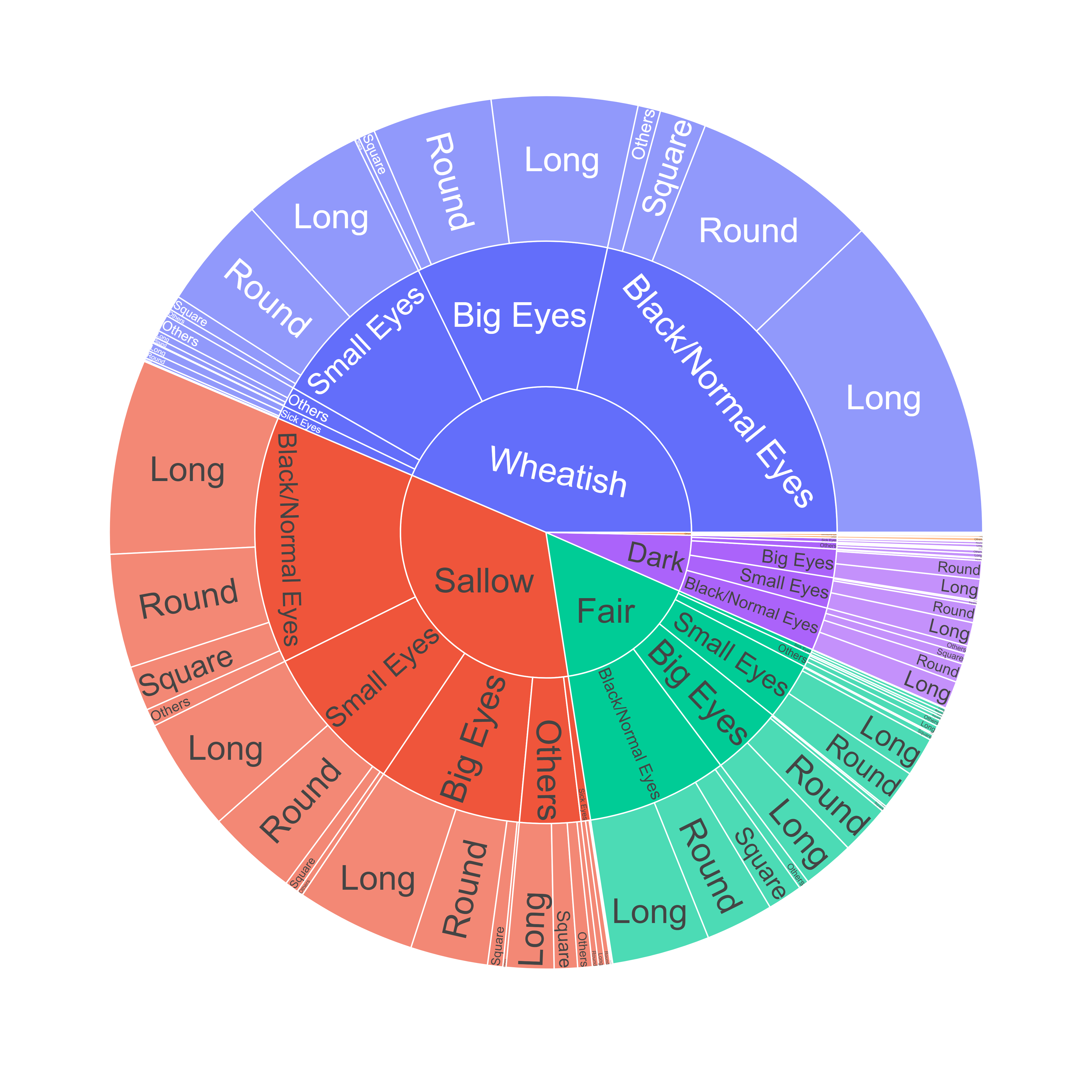}
    \caption*{Selected Dissimilar Images}
  \end{subfigure} 
  }
  
    \subcaptionbox{Complexion}{
      \begin{subfigure}[t]{.35\linewidth}
      \centering
      \includegraphics[width=1.0\textwidth]{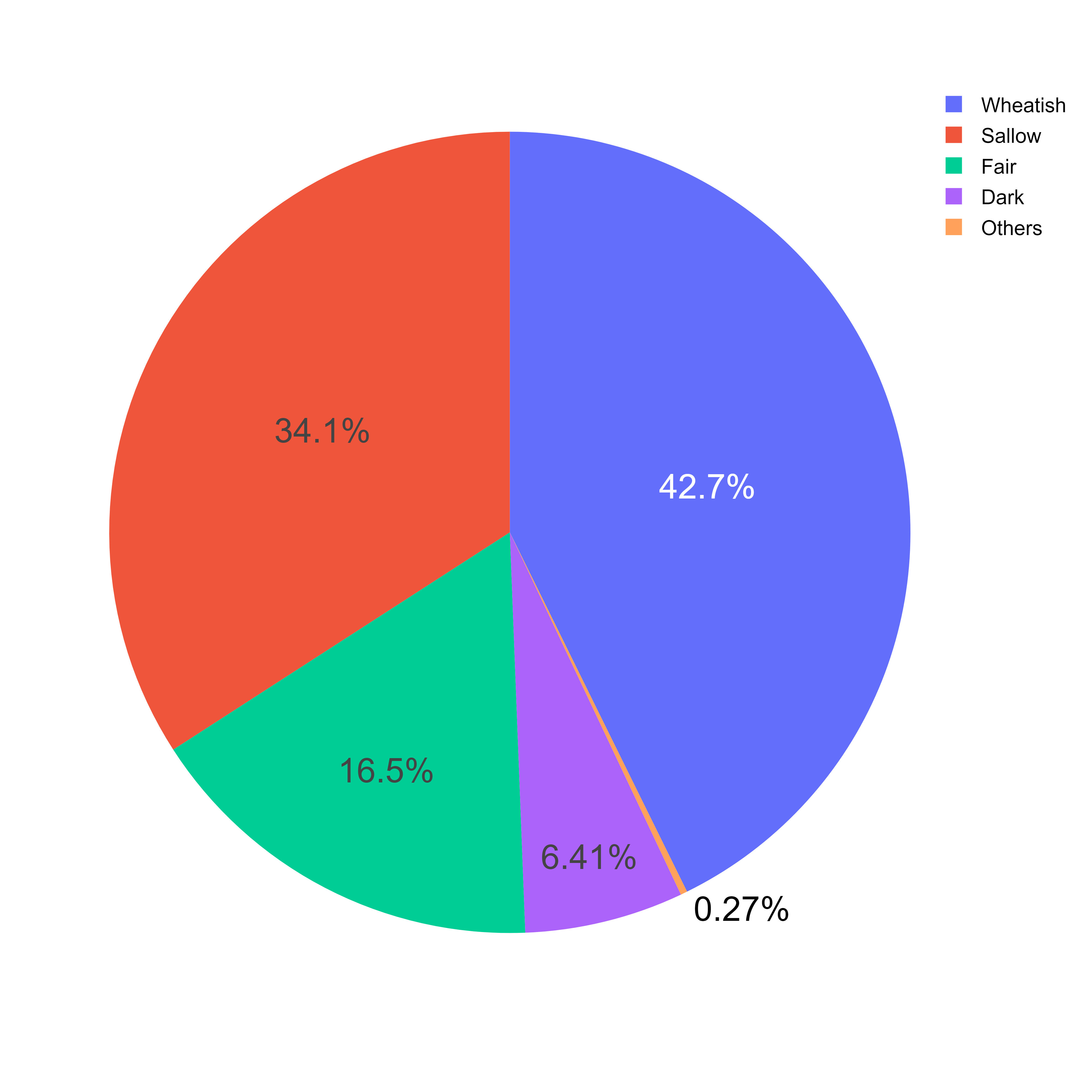}
    \caption*{Full Dataset}
  \end{subfigure}
    \begin{subfigure}[t]{.35\linewidth}
      \centering
      \includegraphics[width=1.0\textwidth]{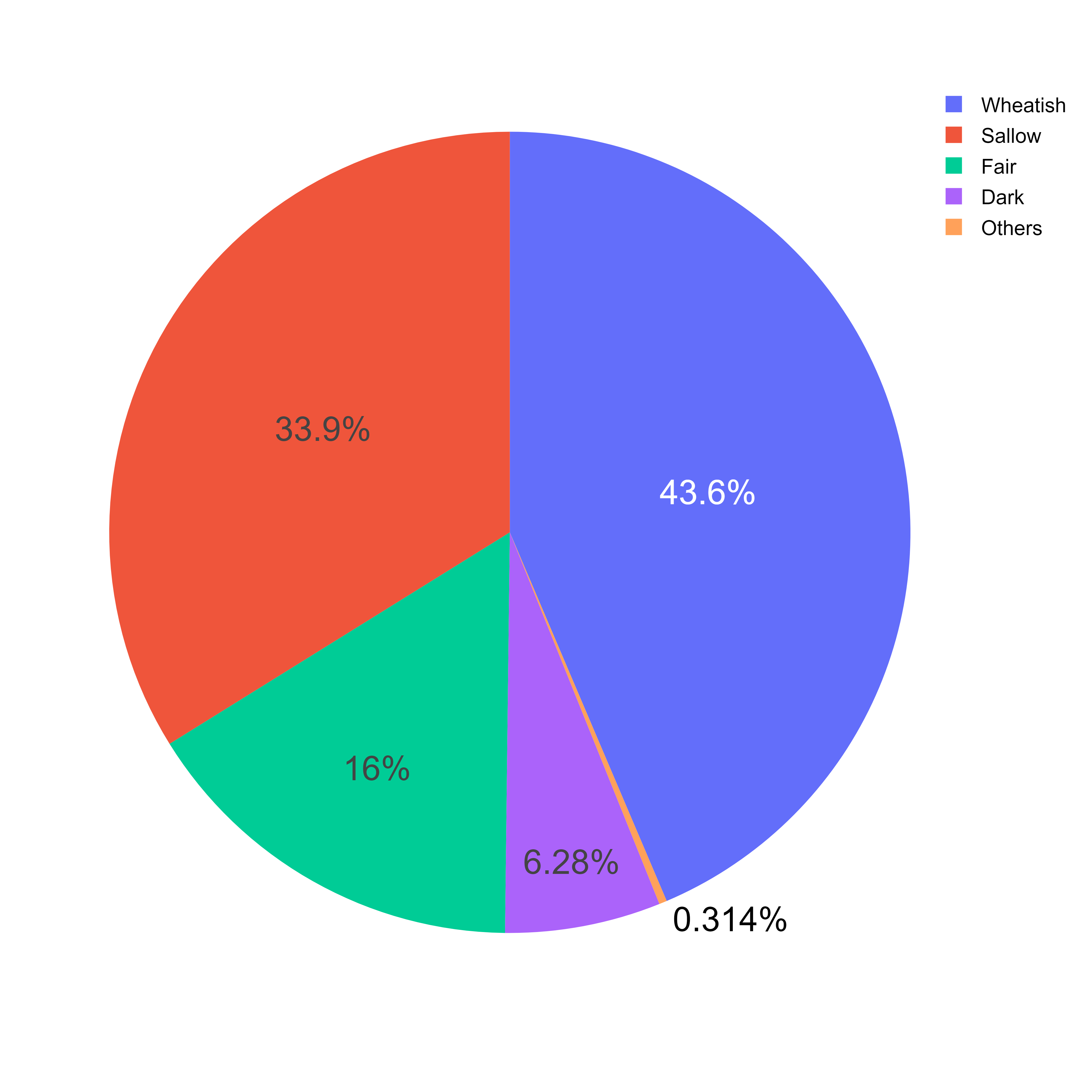}
    \caption*{Selected Dissimilar Images}
  \end{subfigure}
  }
  
      
  \caption{Data distribution between full dataset and dissimilar images generated from the simulation logs with respective to different features}
    \label{fig:distribution_similairty}
\end{figure*}

Figure \ref{fig:distribution_similairty} highlight the data distribution of different sensitive features for the full dataset as well as the dissimilar images obtained from the user simulator logs. As depicted in these plots, both the distribution are similar which supports the claim that the framework employed works in fair manner and is not biased towards specific classes within each tangible factor of variation.

\section{Web Interface}
\label{sec:web_interface}
${\tt FaIRCoP}$ is built on highly scalable and dynamic open-source frameworks: Next.js, Redux, Geist UI (frontend) and Django (backend). Apart from providing an overall theme to the system, advanced techniques such as Dynamic Import, Static HTML Export and Internationalisation have been used to ensure fast loading times and universal deployability. The entire data from visual elements to the results take shape as per variables configured on the backend, which ensures true CMS like behaviour, thus, full control over the frontend. To address security concerns, the app is hosted with it's static HTML export which leaves only a single entry point into the system. After a successful login, a user session is authorized every minute to never leave a user with stale data.

A video demonstration of the working system can be viewed at \href{https://ijcai-faircop-default.layer0-limelight.link/#0}{https://ijcai-faircop-default.layer0-limelight.link/#0}

\section{Future Work}
The system currently starts by showing the user a completely random set of images based on certain attributes explicitly selected by the user. We intend on making this initialization more robust by incorporating components of natural language to improve the system metrics. Furthermore, a metric can be added to this system which determines when to explore the database or when to exploit the previous data, and suggest those images in addition to the base set of recommended images. We also plan to improve the user simulator by using the concept of Eigen faces \cite{eigenfaces} and designing a way to determine visual similarity using its geometric properties.

\bibliographystyle{ACM-Reference-Format}
\bibliography{main}